\documentclass[10pt,letterpaper]{article}
\usepackage{multirow}
\usepackage{wacv}
\usepackage{times}
\usepackage{epsfig}
\usepackage{graphicx}
\usepackage{amsmath}
\usepackage{amssymb}
\usepackage{graphicx}
\usepackage{algorithm}
\usepackage{algpseudocode}
\usepackage{tabularx}
\usepackage{subfig}
\usepackage{wrapfig,lipsum,booktabs}
\usepackage{algorithmicx}
\usepackage{url}
\usepackage{color}

\wacvfinalcopy 

\def\wacvPaperID{183} 

\ifwacvfinal\fi
\begin{document}

\title{Multi-layer Pruning Framework for Compressing Single Shot MultiBox Detector}

\author{Pravendra Singh\\
IIT Kanpur\\
{\tt\small psingh@iitk.ac.in}
\and
Manikandan R \\
R\&D Center, Hitachi India Pvt. Ltd.\\
{\tt\small manikandan@hitachi.co.in}
\and
Neeraj Matiyali \\
IIT Kanpur \\
{\tt\small neermat@iitk.ac.in}
\and
Vinay P. Namboodiri \\
Indian Institute of Technology, Kanpur \\
{\tt\small vinaypn@iitk.ac.in}
}

\maketitle
\ifwacvfinal\thispagestyle{empty}\fi

\begin{abstract}
We propose a framework for compressing state-of-the-art Single Shot MultiBox Detector (SSD). The framework addresses compression in the following stages: Sparsity Induction, Filter Selection, and Filter Pruning. In the Sparsity Induction stage, the object detector model is sparsified via an improved global threshold. In Filter Selection \& Pruning stage, we select and remove filters using sparsity statistics of filter weights in two consecutive convolutional layers. This results in the model with the size smaller than most existing compact architectures. We evaluate the performance of our framework with multiple datasets and compare over multiple methods. Experimental results show that our method achieves state-of-the-art compression of 6.7X and 4.9X on PASCAL VOC dataset on models SSD300 and SSD512 respectively. We further show that the method produces maximum compression of 26X with SSD512 on German Traffic Sign Detection Benchmark (GTSDB). Additionally, we also empirically show our method's adaptability for classification based architecture VGG16 on datasets CIFAR and German Traffic Sign Recognition Benchmark (GTSRB) achieving a compression rate of 125X and 200X with the reduction in flops by 90.50\% and 96.6\% respectively with no loss of accuracy. In addition to this, our method does not require any special libraries or hardware support for the resulting compressed models.
\end{abstract}

\section{Introduction}
Object detection has been a major application that has seen tremendous progress with the advent of deep learning. There have been some techniques in object detection~\cite{Girshick2015FastR,Liu2016SSDSS} that have achieved state of the art accuracy. These advances enable object detection to be widely used for various applications ranging from pedestrian detection for autonomous driving to indoor object detection for robotics applications. One factor that however limits the wide-scale deployment of object detection has been that the models obtained through deep learning are rather large with millions of parameters. It is in this context that research in model compression for object detection is relevant.  However, model compression ~\cite{Han2015DeepCC,Hinton2015DistillingTK} has mostly been researched in the area of object classification. The task of model compression for object detection is more challenging than classification because for classification one relies on the whole image for obtaining the prediction while for detection one need to localize a specific image region that contains the object. Model compression might inaccurately affect specific image regions reducing the object detection performance. In this paper, we show that using a multi-layer pruning algorithm it is possible to obtain significant compression of the model while maintaining the accuracy in detection. The proposed method provides us with state of the art results for model compression for object detection.

In this paper, we propose a method for deep learning compression focusing on the Single Shot MultiBox Detector (SSD). However, it should be noted that the method is general and given a different loss function such as that of classification, the method is still applicable. Further by benchmarking on SSD object detection we can compare our method with other deep model compression techniques that have been proposed for object detection. Our method has the following steps, a sparsity constraint-based training step, a filter selection, and pruning step and finally a retraining step with the resulting selected filters. At the end of the filter selection and pruning step, we observe a drop in accuracy in object detection. However, the retraining step, in the end, ensures that we regain the accuracy in object detection.

Models resulting from our method do not rely on any special hardware/sparse library support. Furthermore, our method is complementary to all other pruning methods such as Binary/Quantized weights, etc.

Through this paper we make the following contributions:

\begin{enumerate}

\item We propose an algorithm for multi-layer pruning with sparsity constraint-based training, filter selection and pruning followed by a retraining step for obtaining a reduced set of weights that provide us with accurate object detection.
\item We investigate our methods over various configurations of SSD across major detection benchmarks such as the PASCAL VOC dataset and the German Traffic Sign Detection Benchmark (GTSDB). Our algorithm performs well for both these benchmark datasets. 
\item We show that this framework is also applicable for object classification. In fact, for the classification task, we obtain a significant reduction of parameters of the order of 125X to 200X. These result in extremely compact models that can be used practically. 

\end{enumerate}

\section{Related work}
Deep neural networks are often over-parameterized with a large number of redundant parameters. Deep compression exploits these redundancies to reduce the overall model size. Multiple approaches are proposed which fall broadly into the following major categories.\\

\noindent
\textbf{Pruning and parameter sharing:}
Initial works that have proposed model compression for classification include optimal brain damage~\cite{LeCun1989OptimalBD} and optimal brain surgeon~\cite{Hassibi1992SecondOD}. Recently there have been some works including work on Deep Compression~\cite{Han2015DeepCC} that uses an iterative threshold based pruning and retraining. Other such methods that have proposed variants of pruning for classification include~\cite{Srinivas2015DatafreePP,Chen2015CompressingNN,Hu2016NetworkTA,Li2016PruningFF}. They vary in the approach they adopt for pruning whether they use a hash function to group weights for pruning~\cite{Chen2015CompressingNN} or use sparsity constraint~\cite{Lebedev2016FastCU}. In our method, we adopt a single multi-layer pruning iteration instead of other techniques that advocate iterative layer pruning strategies. Further, an object detection the loss function objective includes a classification loss function and a localization objective function. This necessitates that any pruning adopted should not be biased by the classification objective alone. In classification pruning may not adversely affect the classification objective as the context present in an image may also help in satisfying the objective. This is not the case with detection. Our approach allows for careful pruning that satisfies both the objective functions.  

\noindent
\textbf{Compact Architectures :} Another set of ideas advocates use of compact architectures. One such approach is that of using transferred convolutional filters motivated from works by Cohen and Welling \cite{Cohen2016GroupEC}, where they use same weights to analyze each part of the image, thereby reducing the number of parameters in the overall model. This approach would be more suited for classes of object detection that have relatively less variance between instances. Objects like people that have high variance cannot be approached using such approaches. Other works~\cite{Li2016MultiBiasNA,Shang2016UnderstandingAI,Zhai2016DoublyCN} focus on methods where they build CNN layer's using multiple base filters. The second set of works look more towards binarization for reducing memory. Binary net \cite{NIPS2016_6573}, BinaryConnect \cite{Courbariaux2015BinaryConnectTD}, XNORNet \cite{Rastegari2016XNORNetIC} focus on training CNN's end to end using binary weights. Compact architectures usually exhibit a high drop in accuracy and are not suitable for object detection.

\noindent
\textbf{Knowledge Distillation:} Knowledge distillation was proposed by Cuarna et al. \cite{Bucila2006ModelC,Ba2014DoDN} and later explored by Hinton et al. \cite{Hinton2015DistillingTK}. Recent works in this area by Balan et al. \cite{Balan2015BayesianDK} used a parametric student model for a Monte Carlo teacher, which was extended to Net2Net \cite{Chen2015Net2NetAL} that accelerated the experimentation process by instantaneously transferring the knowledge from a previous network to each new deeper or wider network. Recently, Zagoruyko et al. \cite{Zagoruyko2016PayingMA} proposed Attention Transfer (AT). They transferred the attention maps that are summaries of the full activations. While the methods described above makes the model thinner and reduce computational cost, it can be only applied to classification tasks with softmax loss function. Our approach is equally applicable for detection as well as classification tasks with the limited trade-off between accuracy and model size.

\noindent
\textbf{Object detection:} While there is a large body of works on compression of classification based architectures, very few works are done till date on compression of object detection based deep learning models. Most notable are works by Xie et al. \cite{Xie2017VisualizationAP} which focuses on compressing SSD by using deconvolution to analyze kernels and active neurons, thereby removing inactive neurons and redundant kernels. Alternatively, Chen et al. \cite{Chen2017LearningEO} proposed a trainable framework for multiclass object detection through knowledge distillation via the introduction of weighed cross entropy and teacher bounded regression for knowledge distillation. Finally, in the very recent work Anisimov et al. \cite{anisimov2017towards}, focus on eliminating channels by sampling using multiple methods and fine-tuning the resulting model on the given task. In conclusion, works that focus on SSD, concentrate either only a few layers of the SSD architecture or use compact base networks which result in reduced performance. In our work, we focus on compressing original SSD with VGG16 as its base network.

\section{Multi-layer Pruning Framework}\label{secFramework}
 
In this section, we give a comprehensive description of our pruning approach. 


\subsection{Framework Overview}
 The core of our pruning framework is based on the principle: ``Evaluate the importance of the filters and remove the unimportant ones" in line with previous works \cite{Luo2017ThiNetAF}.

Let $I_l$ and $W_l$ denote the input tensor and parameters of $l$-th convolutional layer. Here $I_l \in \mathbb{R}^{c_{l-1}\times h_l \times w_l}$ has $c_{l-1}$ channels, $h_l$ rows and $w_l$ columns. The weight tensor $W_l \in \mathbb{R}^{c_l\times c_{l-1}\times k_l \times k_l}$ is a set of $c_l$ filters of $c_{l-1}\times k_l\times k_l$ size each. This convolutional layer produces the output tensor $O_l \in \mathbb{R}^{c_{l}\times h_{l+1} \times w_{l+1}}$, which is a set of $c_l$ feature maps.

Our goal is to remove the unimportant filters in $W_l$. Let ${c'_l\times c_{l-1}\times k_l \times k_l}$ be the new size of $W_l$, where $c'_l$ is the number of remaining filters after pruning of the unimportant ones. Since the number of filters is modified in layer $l$, the size of output tensor $O_l$ and equivalently the size of input tensor of next layer $I_{l+1}$ also reduces from ${c_{l}\times h_{l+1} \times w_{l+1}}$ to ${c'_{l}\times h_{l+1} \times w_{l+1}}$. Hence, the corresponding weights in $W_{l+1}$ also need to be removed, which would in turn reduce the size of $W_{l+1}$ from ${c_{l+1}\times c_{l}\times k_{l+1} \times k_{l+1}}$ to ${c_{l+1}\times c'_{l}\times k_{l+1} \times k_{l+1}}$.

In our multilayer pruning framework, we propose architecture such as SSD in multiple phases. In each phase, we target a different set of consecutive layers. First, we introduce sparsity in these layers and prune them aggressively in a single shot. Then, we recover the performance of the model by fine-tuning it. Unlike previous approaches \cite{Han2015DeepCC,Luo2017ThiNetAF,han2016dsd} where they either do the pruning layer-wise or iteratively prune the full network, our method prunes a set of consecutive conv layers at a time.

\begin{figure}[t]
    \centering
     \scalebox{0.7}{
    \includegraphics[width=\columnwidth]{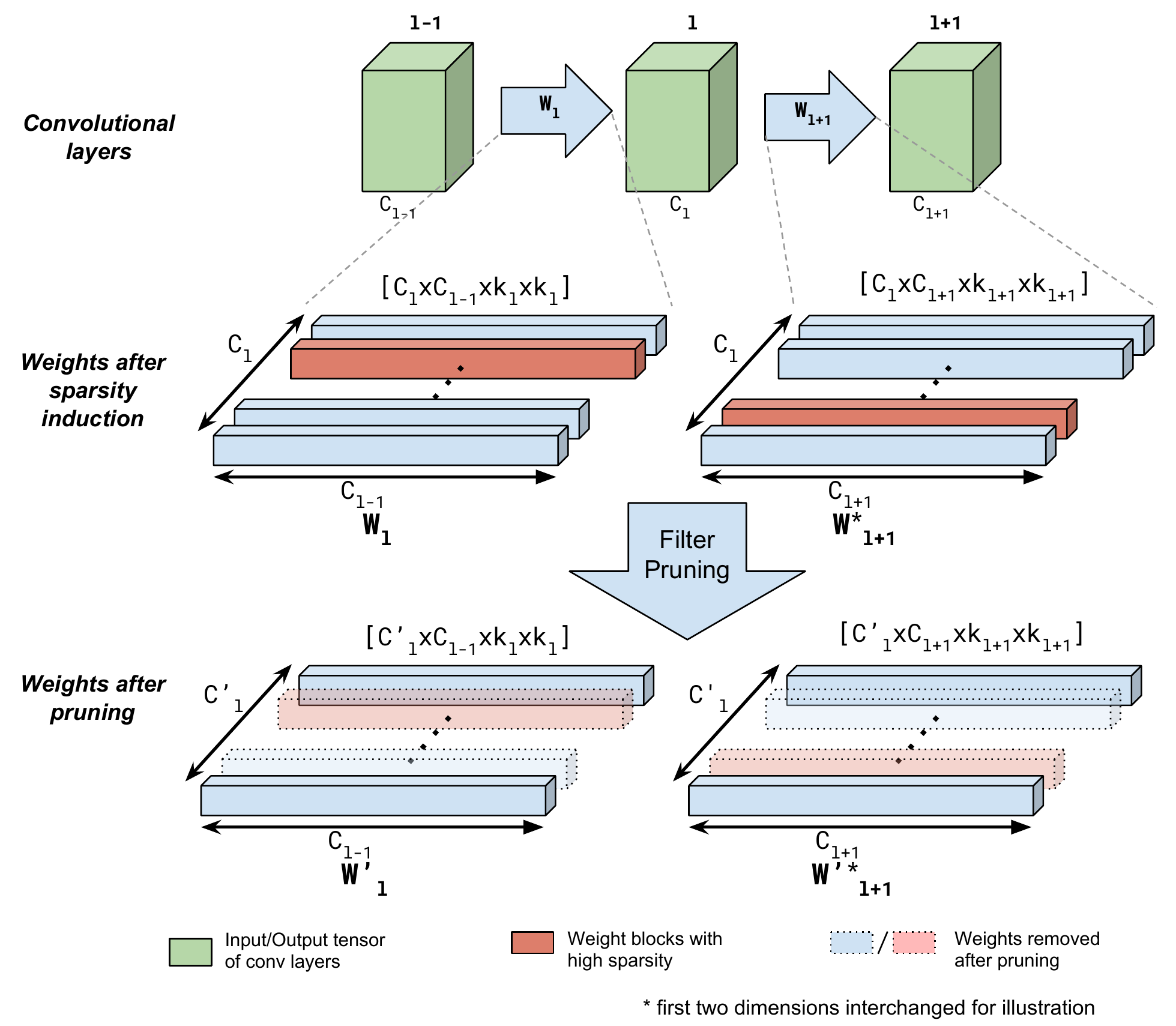}
    }
    \caption{Illustration of filter selection and pruning of convolutional filters in our pruning framework. After introducing sparsity in network weights by training with L1 regularization and thresholding, we examine the filter weights $W_l$ and $W_{l+1}$ of two consecutive convolutional layers $l$ and $l+1$ respectively. Based on sparsity in these weights, we select the filters in $W_l$ to be pruned. Note that corresponding weights in $W_{l+1}$ also get removed. Removing these weights reduces the number of output channels in $l$-th layer from $c_l$ to $c_l'$. }
  
    \label{fig:pruning}
\end{figure}

Figure~\ref{fig:pruning} shows a schematic of our filter pruning strategy. Given a pre-trained model with parameters $\Theta$ and a set of consecutive layers $L$ to be pruned. The entire framework executes in the following steps.

\begin{enumerate}
\item \textbf{Sparsity induction:} In the first step, we train the model with loss-function with L1-norm resulting in the sparse model $\Theta_{L1}$. Next, we set all the weights in $L$ with an absolute value smaller than a threshold to zero resulting in the model $\Theta_{L1}^{th}$. The value of the threshold is chosen based on the performance of $\Theta_{L1}^{th}$ on the validation set. Unlike previous approaches \cite{han2016dsd} that select layer-wise thresholds, we use a single global threshold for the whole layer set $L$. This allows us to prune filters in $L$ in a single step. 
We explain this in detail in Section~\ref{secSpInd}.

\item \textbf{Filter selection:} To evaluate the importance of a filter in layer $l\in L$, we use filter sparsity statistics of layer $l$ and $l+1$ in model $\Theta_{L1}^{th}$. Unlike previous approaches that use data-driven methods to figure out the importance, we use only weight statistics. The key idea here is to remove all the filters in $l$ that have a large fraction of zero weights in them as well as those filters in $l$ that have a large fraction of zero weights corresponding to them in the filters of the following layer $l+1$ as illustrated in Figure~\ref{fig:pruning}. At the end of this step, we obtain a list of filters in layer $l$ that are deemed as removable from the model. We repeat step 2 for each layer in $L$. A more detailed explanation is presented in Section~\ref{secFilPr}.

\item \textbf{Pruning:} Filters selected in step 2 for layer $l$ and corresponding weights in layer $l+1$ associated with the output of these filters are removed from model $\Theta_{L1}$. This is repeated for each layer in $L$ sequentially.

\item \textbf{Retraining:} Finally, we retrain the pruned network using the original loss (without L1 regularization) to recover the performance drop due to sparsity induction in step 1.
\end{enumerate}

\subsection{Set-wise optimized solution}

Instead of finding a layer-wise optimized solution (layer-wise thresholding), we are finding the set-wise optimized solution (global thresholding for the whole layer-set) which results in high pruning with less performance drop because it preserves dependency across layers in the layer-set $L$.


\subsection{Sparsity Induction}\label{secSpInd}

Let $D=\left\{(x_0,y_0),(x_1,y_2),\ldots,(x_n,y_n)\right\}$ be the training set, where $x_i$ and $y_i$ are input and target label. The parameters $\Theta$ of the original model are optimized to minimize the cost $C_D(\Theta)$.
\begin{equation}
\Theta =  \textrm{argmin} \{C_D(\Theta)\}
\end{equation}
The form of this cost function depends on the task to be solved by the original network. For instance, we use multibox loss function for SSD.

Let $L=\{l_{\textrm{start}}, \ldots, l_{\textrm{end}}\}$ be the set of consecutive layers to be pruned. In order to induce sparsity in parameters of layers in $L$, we add L1-norm of parameters of these layers to the original cost function and train the network initialised with $\Theta$.
\begin{equation}
\Theta_{L1} =  \textrm{argmin} \left\{C_D(\Theta) + \alpha \sum_{l=l_{\textrm{start}}}^{l_{\textrm{end}}}||W_l||_1\right\}
\end{equation}

We choose the regularisation factor $\alpha$ such that the performance of the model with new parameters on a validation set $D_\text{val}$ is close to the original performance, i.e. $P_{D_\text{val}}(\Theta_{L1}) \ge P_{D_\text{val}}(\Theta) - \epsilon_1$. Here, $\epsilon_1$ is the tolerance constant that allows us to control the degree of sparsity in the parameters of layers in $L$. $P_{D_\text{val}}(\Theta)$ is the performance (eg. accuracy, AP etc.) of model with parameters $\Theta$ on the validation set $D_\text{val}$. 

After obtaining $\Theta_{L1}$, we set all the weights in $L$ with absolute values smaller than a threshold $t$ to zero. This gives us the parameters $\Theta_{L1}^{th}$. We search for optimal $t$ in a range proportional to standard deviation of weights in $L$, such that  $P_{D_\text{val}}(\Theta_{L1}^{th}) \ge P_{D_\text{val}}(\Theta_{L1}) - \epsilon_2$. The constant $\epsilon_2$ provides us the additional control on the number of zero weights in $L$. At this point, our network parameters are ready for filter pruning in layers in $L$, which is described in next section.

\subsection{Filter pruning}\label{secFilPr}

\begin{algorithm}[t] \caption{Algorithm for filter pruning}\label{algoFilPru}

\begin{algorithmic}
 
\State {\bf Inputs}: Parameters of network after sparsity induction, $\Theta_{L1}$ and $\Theta_{L1}^{th}$; set of layers to be pruned $L$.
\State {\bf Output}: : New compressed model $M_c$ with weights $\Theta_c$.
\State filterindex = empty-list();
\State $\Theta_c$ = $\Theta_{L1}$
\For {layer $l$ in  $L$}
    \For {filter $i$ in $\{1, \ldots, c_l\}$}
    \State Extract $F_i$ and $G_i$ from $\Theta_L^{th}$
    \State splevelF = Sparsity-level($F_i$);
    \State splevelG = Sparsity-level($G_i$);
    \If{splevelF $>=s'_F$} 
          \If {splevelF $>=s_F$}   
               \State filterindex.add$(i)$
          \Else
               \If{splevelG $>=s_G$}
                       \State filterindex.add$(i)$
                \EndIf
           \EndIf   
     \EndIf
     \EndFor
     \For {filter $i$ in filterindex}
     \State $\Theta_c$ = Remove $F_i$ from $W_l$, $G_i$ from $W_{l+1}$ in $\Theta_{c}$
      \State $\Theta_L^{th}$ = Remove $F_i$ from $W_l$, $G_i$ from $W_{l+1}$ in $\Theta_L^{th}$
     \EndFor
\EndFor
\State $M_c$ = Redefine model with the remaining parameters $\Theta_{c}$
\State Initialize $M_c$ with $\Theta_{c}$
\State Return Model $M_c$ and $\Theta_{c}$

\end{algorithmic}
\end{algorithm}
In this step, we determine filter importance corresponding to layer $l\in L$ . This is executed by examining sparsity statistics of $W_l$ and $W_{l+1}$ in $\Theta_{L1}^{th}$. Let $F_i \in \mathbb{R}^{c_{l-1}\times k_l\times k_l}$ be the $i$-th of $c_l$ filters in $W_l$. Note that in the next layer, the output feature map of filter $F_i$ is connected to a slice of tensor $W_{l+1}$ of size ${c_{l+1}\times 1\times k_{l+1}\times k_{l+1}}$, we call  this slice of weights $G_i$. Essentially, our filter pruning strategy is to remove the $i$-th filter if either one of the tensors $F_i$ and $G_i$ has a very large fraction of zero weights (see Figure~\ref{fig:pruning}). We implement this strategy using three thresholds $s_F$, $s'_F$ and $s_G$ with $s_F>s'_F$. For each $i\in \{1, \ldots, c_l\}$, we evaluate following conditions and remove the $i$-th if atleast one of them holds true:
\begin{enumerate}
\item sparsity statistics in $F_i$ is higher than $s_F$

\item sparsity statistics in $F_i$ is higher than $s'_F$ and sparsity statistics in $G_i$ is higher than $s_G$.

\end{enumerate}

Intuitively, the first condition selects the filters that have a very high level of sparsity ($>s_F$). These filters can be safely removed because their output activation maps are very weak. The second condition further selects those filters whose output activations are stronger than those selected by the first condition ($s_F > $ sparsity level $>s_F'$) but have an overall low contribution because of high sparsity in the weights connected to them in the next layer (sparsity level $>s_G$). Our method for computing sparsity statistics is unlike previous works \cite{Hu2016NetworkTA} where the method is data-driven and measures sparsity on layers activation (e.g. output of ReLU) rather than its filter weights. To compute the sparsity level, we use the concept of zero row. A row in a filter $k_l\times k_l$ is considered as a zero row (size $1\times k_l$) if all elements in the row are zero. Sparsity level is simply the percentage of zero rows in a filter.
\begin{equation*}
\text{Sparsity level}(F_i) = \dfrac{\text{Number of zero rows}}{c_{l-1}\times k_l}
\end{equation*}
\begin{equation*}
\text{Sparsity level}(G_i) = \dfrac{\text{Number of zero rows}}{c_{l+1}\times k_{l+1}} 
\end{equation*}

Let $\Theta_c$ be the parameters of the network after filter pruning. We empirically found that taking the values of parameters in $\Theta_c$ from $\Theta_{L1}$ rather than from $\Theta_{L1}^{th}$ results in improved performance. The reason for improved performance is due to the restoration of values of zero weights that did not get pruned. $\Theta_{L1}^{th}$ only serve as a guide for the selection of filters to be pruned. Algorithm~\ref{algoFilPru} describes the full procedure of filter pruning in detail.

Finally, we retrain the pruned network without L1 regularization to restore the performance drop due to sparsity induction.


\section{Experiments}

We apply our multilayer pruning framework to compress Single Shot MultiBox Detector (SSD). We experiment on both smaller and larger variant of SSD, namely SSD300 and SSD512. In our initial experiments on German Traffic Sign Detection Benchmark (GTSDB) \cite{Houben-IJCNN-2013}, we achieved a compression of SSD512 by a factor of 26X. To further test the effectiveness of our multi-layer pruning framework, we run experiments on PASCAL VOC \cite{Everingham2014ThePV} where we compress SSD300 by a factor of 7X. We also compare our pruned SSD models with other existing light-weight SSD variants, in terms of model size and parameters. We perform additional experiments on the COCO object detection set \cite{cocomicrosoft}. On COCO, we compress SSD300 by a factor of $ \sim $ 3X with no loss of AP@0.5. Please refer to the appendix for more details on our experiments on COCO.

To test the effectiveness of our method and its task independence across the convolutional neural network, we perform experiments for pruning of image classification architecture namely VGG16 on datasets German Traffic Sign Recognition Benchmark (GTSRB) \cite{Houben-IJCNN-2013} and CIFAR10. On CIFAR10, we compress the VGG16 by a factor 125X with 90.5\% reduction in flops.

\subsection{Pruning of SSD on PASCAL VOC}\label{secExpVoc}

\begin{figure}[t]
    \centering
     
        \subfloat[conv3\_3]{\includegraphics[width=0.45\columnwidth]{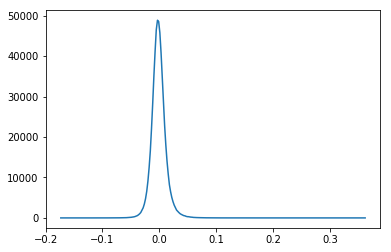}}
        \quad
        \subfloat[conv4\_3]{\includegraphics[width=0.45\columnwidth]{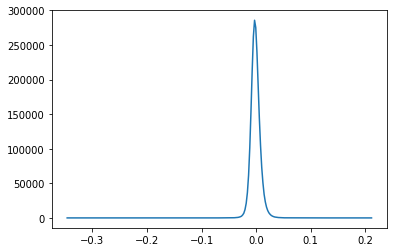}}
        \\                
        \subfloat[conv3\_3 (with L1)]{\includegraphics[width=0.45\columnwidth]{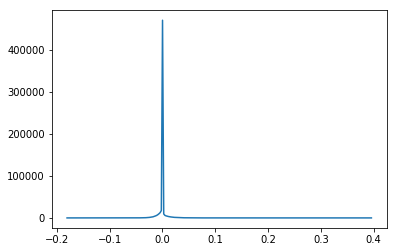}}
        \quad
        \subfloat[conv4\_3 (with L1)]{\includegraphics[width=0.45\columnwidth]{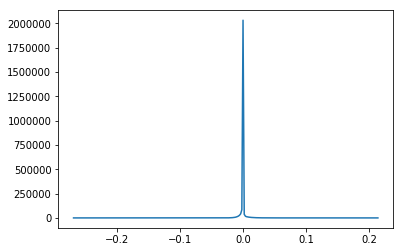}}
       
    \caption{Weight distributions of convolutional layers conv3\_3 and conv4\_3 of SSD network without (first row) and with (second row) L1 regularization in phase 1.}

    \label{figssdhist}
\end{figure}

\begin{figure}[ht!]
\centering
\includegraphics[width=0.48\linewidth]{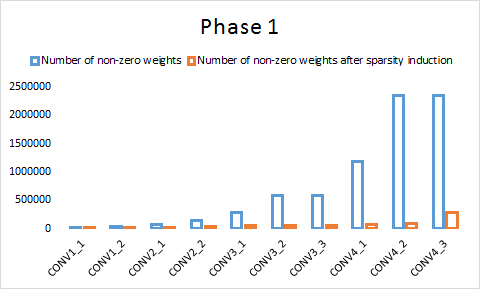}~~
\includegraphics[width=0.48\linewidth]{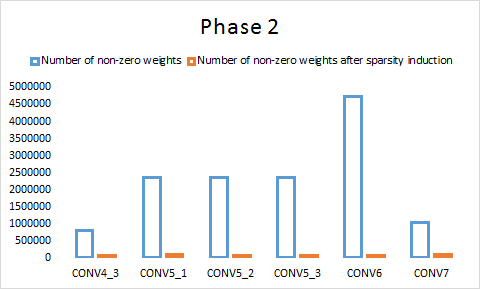}\\
\includegraphics[width=0.48\linewidth]{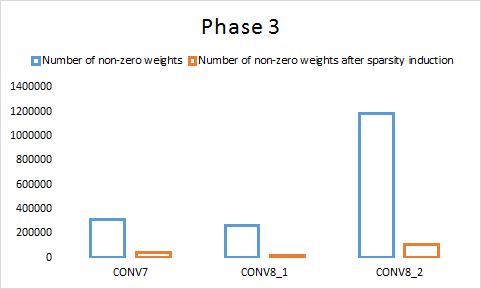}~~
\includegraphics[width=0.48\linewidth]{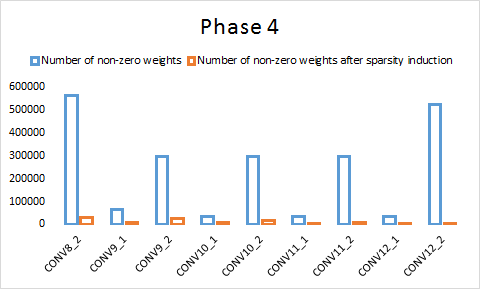}
\caption{Number of non-zero parameters in SSD convolutional layers before (blue) and after (orange) different phases of our sparsity induction method on PASCAL VOC dataset.}

\label{figssdsparse}
\end{figure}

\begin{table*}[t]
\centering
\scalebox{0.72}{
\begin{tabular}{|c|c|c|c|c|c|c|c|}
\hline
\multicolumn{2}{|c|}{\multirow{2}{*}{\textbf{}}} & \multicolumn{3}{c|}{\textbf{SSD300 on  PASCAL VOC}} & \multicolumn{3}{c|}{\textbf{SSD512 on  PASCAL VOC}} \\ \cline{3-8} 
\multicolumn{2}{|c|}{} & \textbf{\texttt{original}} & \textbf{\texttt{pruned-3det}} & \textbf{\texttt{pruned-6det}} & \textbf{\texttt{original}} & \textbf{\texttt{pruned-4det}} & \textbf{\texttt{pruned-7det}} \\ \hline
\multicolumn{2}{|c|}{\textbf{Input Size}} & 300X300X3 & 300X300X3 & 300X300X3 & 512X512X3 & 512X512X3 & 512X512X3 \\ \hline
\multirow{25}{*}{\textbf{Layers}} & \textbf{CONV1\_1} & 3X3 CONV, 64 & 3X3 CONV, 64 & 3X3 CONV, 64 & 3X3 CONV, 64 & 3X3 CONV, 64 & 3X3 CONV, 64 \\ \cline{2-8} 
 & \textbf{CONV1\_2} & 3X3 CONV, 64 & 3X3 CONV, 56 & 3X3 CONV, 56 & 3X3 CONV, 64 & 3X3 CONV, 61 & 3X3 CONV, 61 \\ \cline{2-8} 
 & \textbf{CONV2\_1} & 3X3 CONV, 128 & 3X3 CONV, 107 & 3X3 CONV, 107 & 3X3 CONV, 128 & 3X3 CONV, 119 & 3X3 CONV, 119 \\ \cline{2-8} 
 & \textbf{CONV2\_2} & 3X3 CONV, 128 & 3X3 CONV, 121 & 3X3 CONV, 121 & 3X3 CONV, 128 & 3X3 CONV, 122 & 3X3 CONV, 122 \\ \cline{2-8} 
 & \textbf{CONV3\_1} & 3X3 CONV, 256 & 3X3 CONV, 193 & 3X3 CONV, 193 & 3X3 CONV, 256 & 3X3 CONV, 215 & 3X3 CONV, 215 \\ \cline{2-8} 
 & \textbf{CONV3\_2} & 3X3 CONV, 256 & 3X3 CONV, 158 & 3X3 CONV, 158 & 3X3 CONV, 256 & 3X3 CONV, 160 & 3X3 CONV, 160 \\ \cline{2-8} 
 & \textbf{CONV3\_3} & 3X3 CONV, 256 & 3X3 CONV, 195 & 3X3 CONV, 195 & 3X3 CONV, 256 & 3X3 CONV, 187 & 3X3 CONV, 187 \\ \cline{2-8} 
 & \textbf{CONV4\_1} & 3X3 CONV, 512 & 3X3 CONV, 263 & 3X3 CONV, 263 & 3X3 CONV, 512 & 3X3 CONV, 252 & 3X3 CONV, 252 \\ \cline{2-8} 
 & \textbf{CONV4\_2} & 3X3 CONV, 512 & 3X3 CONV, 181 & 3X3 CONV, 181 & 3X3 CONV, 512 & 3X3 CONV, 172 & 3X3 CONV, 172 \\ \cline{2-8} 
 & \textbf{CONV4\_3  D1} & 3X3 CONV, 512 & 3X3 CONV, 331 & 3X3 CONV, 331 & 3X3 CONV, 512 & 3X3 CONV, 313 & 3X3 CONV, 313 \\ \cline{2-8} 
 & \textbf{CONV5\_1} & 3X3 CONV, 512 & 3X3 CONV, 98 & 3X3 CONV, 98 & 3X3 CONV, 512 & 3X3 CONV, 170 & 3X3 CONV, 170 \\ \cline{2-8} 
 & \textbf{CONV5\_2} & 3X3 CONV, 512 & 3X3 CONV, 108 & 3X3 CONV, 108 & 3X3 CONV, 512 & 3X3 CONV, 134 & 3X3 CONV, 134 \\ \cline{2-8} 
 & \textbf{CONV5\_3} & 3X3 CONV, 512 & 3X3 CONV, 78 & 3X3 CONV, 78 & 3X3 CONV, 512 & 3X3 CONV, 117 & 3X3 CONV, 117 \\ \cline{2-8} 
 & \textbf{CONV6} & 3X3 CONV, 1024 & 3X3 CONV, 146 & 3X3 CONV, 146 & 3X3 CONV, 1024 & 3X3 CONV, 305 & 3X3 CONV, 305 \\ \cline{2-8} 
 & \textbf{CONV7  D2} & 1X1 CONV, 1024 & 1X1 CONV, 106 & 1X1 CONV, 106 & 1X1 CONV, 1024 & 1X1 CONV, 332 & 1X1 CONV, 332 \\ \cline{2-8} 
 & \textbf{CONV8\_1} & 1X1 CONV, 256 & 1X1 CONV, 34 & 1X1 CONV, 34 & 1X1 CONV, 256 & 1X1 CONV, 122 & 1X1 CONV, 122 \\ \cline{2-8} 
 & \textbf{CONV8\_2  D3} & 3X3 CONV, 512 & 3X3 CONV, 198 & 3X3 CONV, 198 & 3X3 CONV, 512 & 3X3 CONV, 133 & 3X3 CONV, 133 \\ \cline{2-8} 
 & \textbf{CONV9\_1} & 1X1 CONV, 128 &  & 1X1 CONV, 12 & 1X1 CONV, 128 & 1X1 CONV, 89 & 1X1 CONV, 89 \\ \cline{2-8} 
 & \textbf{CONV9\_2  D4} & 3X3 CONV, 256 &  & 3X3 CONV, 36 & 3X3 CONV, 256 & 3X3 CONV, 168 & 3X3 CONV, 168 \\ \cline{2-8} 
 & \textbf{CONV10\_1} & 1X1 CONV, 128 &  & 1X1 CONV, 22 & 1X1 CONV, 128 &  & 1X1 CONV, 81 \\ \cline{2-8} 
 & \textbf{CONV10\_2  D5} & 3X3 CONV, 256 &  & 3X3 CONV, 41 & 3X3 CONV, 256 &  & 3X3 CONV, 92 \\ \cline{2-8} 
 & \textbf{CONV11\_1} & 1X1 CONV, 128 &  & 1X1 CONV, 22 & 1X1 CONV, 128 &  & 1X1 CONV, 40 \\ \cline{2-8} 
 & \textbf{CONV11\_2  D6} & 3X3 CONV, 256 &  & 3X3 CONV, 66 & 3X3 CONV, 256 &  & 3X3 CONV, 84 \\ \cline{2-8} 
 & \textbf{CONV12\_1} &  &  &  & 1X1 CONV, 128 &  & 1X1 CONV, 40 \\ \cline{2-8} 
 & \textbf{CONV12\_2  D7} &  &  &  & 4X4 CONV, 256 &  & 4X4 CONV, 80 \\ \hline
\multicolumn{2}{|c|}{\multirow{2}{*}{\textbf{Total Parameters}}} & \multirow{2}{*}{26.3M} & \multirow{2}{*}{3.7M (14.1\%)} & \multirow{2}{*}{3.9M (14.8\%)} & \multirow{2}{*}{27.2M} & \multirow{2}{*}{5.1M (18.8\%)} & \multirow{2}{*}{5.5M (20.2\%)} \\
\multicolumn{2}{|c|}{} &  &  &  &  &  &  \\ \hline
\multicolumn{2}{|c|}{\multirow{2}{*}{\textbf{Model Size}}} & \multirow{2}{*}{105.2 MB} & \multirow{2}{*}{15 MB (7X)} & \multirow{2}{*}{15.7 MB (6.7X)} & \multirow{2}{*}{108.8 MB} & \multirow{2}{*}{20.3 MB (5.4X)} & \multirow{2}{*}{22 MB (4.9X)} \\
\multicolumn{2}{|c|}{} &  &  &  &  &  &  \\ \hline
\multicolumn{2}{|c|}{\textbf{Mean AP}} & 77.16 & 72.15 & 75.07 & 79.52 & 75.59 & 77.94 
\\ \hline
\end{tabular}
}
\caption{Layer-wise pruning results in SSD300 and SSD512 models on PASCAL VOC dataset. CONVX\_DN  denotes the N-th layer that is connected to detection layer (conf and loc layer).}

\label{tabssdlayer}
\end{table*}

\begin{table*}[t]
\centering
\scalebox{0.5}{
\begin{tabular}{|c|c|c|c|c|c|c|c|c|c|c|c|c|c|c|c|c|c|c|c|c|c|c|}
\hline
\multicolumn{1}{|c|}{\textbf{Method}} & \textbf{data} & \textbf{mAP} & \textbf{aero} & \textbf{bike} & \textbf{bird} & \textbf{boat} & \textbf{bottle} & \textbf{bus} & \textbf{car} & \textbf{cat} & \textbf{chair} & \textbf{cow} & \textbf{table} & \textbf{dog} & \textbf{horse} & \textbf{mbike} & \textbf{person} & \textbf{plant} & \textbf{sheep} & \textbf{sofa} & \textbf{train} & \textbf{tv} \\ \hline
\textbf{\texttt{SSD512-original}} & 07+12 & 79.52 & 86.82 & 85.3 & 79.21 & 71.97 & 56.51 & 86.3 & 88.02 & 88.83 & 63.04 & 86.61 & 75.16 & 86.33 & 87.64 & 86.26 & 80.83 & 52.73 & 84.02 & 78.54 & 88 & 78.29 \\ \hline
\textbf{\texttt{SSD512-pruned-4det}} & 07+12 & 75.59 & 84.76 & 82.87 & 76.29 & 69.42 & 55.13 & 84.5 & 85.52 & 82.62 & 62.29 & 83.06 & 67.89 & 77.15 & 83.16 & 82.22 & 79.25 & 50.7 & 78.02 & 68.46 & 82.4 & 76.06 \\ \hline
\textbf{\texttt{SSD512-pruned-7det}} & 07+12 & 77.94 & 85.32 & 85.17 & 76.25 & 71.77 & 57.24 & 85.91 & 87.22 & 86.9 & 63.38 & 84.55 & 75.15 & 81.95 & 85.95 & 83.76 & 80.47 & 51.67 & 77.79 & 76.09 & 85.97 & 76.29 \\ \hline
\textbf{\texttt{SSD300-original}} & 07+12 & 77.16 & 80.4 & 82.95 & 74.62 & 71.61 & 50.49 & 86.04 & 86.55 & 88.02 & 60.88 & 83.1 & 77.87 & 85.55 & 86.68 & 84.14 & 78.26 & 50.44 & 74.28 & 80.03 & 85.88 & 75.49 \\ \hline
\textbf{\texttt{SSD300-pruned-3det}} & 07+12 & 72.15 & 76.85 & 82.04 & 71.59 & 63.8 & 49.28 & 79.73 & 82.94 & 78.32 & 58.88 & 76.5 & 63.28 & 74.59 & 80.81 & 78.66 & 77.02 & 47.79 & 75.24 & 69.21 & 82.02 & 74.46 \\ \hline
\textbf{\texttt{SSD300-pruned-6det}} & 07+12 & 75.07 & 78.01 & 82.45 & 72.99 & 67.05 & 48.83 & 82.64 & 86.08 & 86 & 60.07 & 76.35 & 74.36 & 82.22 & 83.44 & 83.44 & 78.54 & 48.45 & 74 & 77.67 & 84.21 & 74.68 \\ \hline
\end{tabular}
}
\caption{AP for each class with original SSD and pruned SSD models on VOC2007 test dataset. Training data, 07+12 is the union of the VOC2007 and VOC2012 trainval dataset.}
\label{tabssdvoc}
\end{table*}

SSD consists of a large number of convolutional layers, where multiple convolutional layers contribute to the detections at various scales. The complex dependency of output detections on convolutional layers makes it challenging to prune. Previous works \cite{Xie2017VisualizationAP} only attempt to prune the base network of SSD.  In our experiments, we prune the full network including the detection layers. However, we observed that pruning the entire network together in a single step massively degrades the performance. Therefore, we apply our multilayer pruning strategy and target the network in chunks of layers in multiple phases. 

In our experiments, we use SSD with base network VGG16. For pruning of SSD512, we employ our method in 4 phases: conv1\_1 - conv4\_3, conv4\_3 - conv7, conv7 - conv8\_2 and conv8\_2 - conv12\_2. We also include the conv layers that perform the localization and classification in these groups of layers.

In each phase, we introduce sparsity in the model and prune the filters of conv layers as discussed in Section~\ref{secFramework}. Figure~\ref{figssdhist} shows the effect of L1 regularization on weight distributions of conv layers in phase 1. Note that the weights have become massively sparse. After thresholding, the number of non-zero parameters in conv layers reduces significantly as shown in Figure~\ref{figssdsparse}. After pruning the filters in four phases, we test two pruned networks on VOC 2007 test set: SSD512-pruned-4det and SSD512-pruned-7det with four and seven detection layers respectively. Starting with 79.52\% mAP of the original model, we achieve mAP 75.59\% and 77.94\% and a reduction of 81.2\% and 79.8\% in number of parameters with SSD512-pruned-4det and SSD512-pruned-7det respectively. This results in 5.4X and 4.9X reduction in model size. Table~\ref{tabssdlayer} shows number of filters and performance of original and pruned models on PASCAL VOC. 
\\
\begin{table}[t]
\centering
\scalebox{1.0}{
\begin{tabular}{| c | c | c |}
\hline
\textbf{Model} & \textbf{AP} & \textbf{MParams} \\ \hline
SqueezeNet1.0\_SSD \cite{anisimov2017towards,iandola2016squeezenet} & 38.45 & 7 \\ \hline
ResNet10\_SSD \cite{anisimov2017towards,he2016deep} & 64.83 & 6.7 \\ \hline
PVANet\_SSD \cite{anisimov2017towards, kim2016pvanet} & 67.69 & 8.1 \\ \hline
MobileNet\_SSD \cite{anisimov2017towards,howard2017mobilenets}& 70.04 & 8.8 \\ \hline
SSDM\_7.5 \cite{anisimov2017towards} & 73.08 & 10.1 \\ \hline
\textbf{SSD300-pruned-3det(ours)} & 72.15 & 3.7 \\ \hline
\textbf{SSD300-pruned-6det(ours)} & \textbf{75.07} & \textbf{3.9} \\ \hline
\textbf{SSD512-pruned-4det(ours)} & \textbf{75.59} & \textbf{5.1} \\ \hline
\textbf{SSD512-pruned-7det(ours)} & \textbf{77.94} & \textbf{5.5} \\ \hline
\end{tabular}}
\caption{Comparison of our pruned SSD models with other existing light-weight variants of SSD on VOC 07 test dataset.}
\label{tabssdparams}
\end{table}

For SSD300 architecture, we prune the model in 4 phases: conv1\_1 - conv4\_3, conv4\_3 - conv7, conv7 - conv8\_2 and conv8\_2 - conv11\_2. We test two pruned networks on VOC 2007 test set: SSD300-pruned-3det and SSD300-pruned-6det with three and six detection layers respectively as shown in Table~\ref{tabssdlayer}. AP for each class with original SSD and pruned SSD models on VOC2007 test dataset is shown in Table~\ref{tabssdvoc}. Table~\ref{tabssdparams} shows the comparison of performance and model parameters (in Millions) of our pruned SSD models with other existing light-weight SSD variants. We also compare SSD300-pruned-6det with compressed F-RCNN models of \cite{he2017channel} and found that our pruned model has more than $34X$ lesser parameters than theirs as shown in Table~\ref{tabsln}(a).

\subsubsection{Baselines}

To the best of our knowledge, this is the first work that compresses all layers of SSD including the base network and the detection layers. Therefore, we have created a baseline to show the efficacy of our method. Our baseline RRF-x-6det is obtained by randomly removing x\% filters from all layers of SSD300 up to CONV7. Please refer to Table~\ref{tabsln}(b) for comparison with the baseline. 

\begin{table*}[t]
\centering
\scalebox{0.92}{
\begin{tabular}{|c|c|c|c|c|}
\hline
\textbf{Methods}     & \textbf{\#Params} & \textbf{Model Size} & \textbf{mAP}  & \textbf{FPS} \\ \hline
F-RCNN 2X \cite{he2017channel}   & 133M              & 509 MB              & 68.3    & -      \\ \hline
F-RCNN 4X \cite{he2017channel}    & 130M              & 500 MB              & 66.9   & 11        \\ \hline
\textbf{6det (ours)} & \textbf{3.9M}     & \textbf{15.7 MB}    & \textbf{75.07}   & \textbf{142} \\ \hline
\end{tabular}
\begin{tabular}{|c|c|c|c|c|}
\hline
\textbf{Methods}     & \textbf{data} & \textbf{\#Params} & \textbf{Model Size} & \textbf{mAP}   \\ \hline
RRF-0.5-6det             & 07+12         & 9.87M             & 39.5 MB             & 73.2           \\ \hline
RRF-0.65-6det            & 07+12         & 7M                & 27.9 MB             & 63.3           \\ \hline
\textbf{6det (ours)} & 07+12         & \textbf{3.9M}     & \textbf{15.7 MB}    & \textbf{75.07} \\ \hline
\end{tabular}}

\caption{ Comparison of our pruned model SSD300-pruned-6det with (from left to right) (a) the state-of-the-art filter pruning approach (ICCV'17) \cite{he2017channel} proposed for Faster R-CNN (b) the baseline Randomly Removing Half Filters (RRF-0.5) and Randomly Removing 65\% Filters (RRF-0.65) from SSD300 (up to CONV7, 6 Detection Layer) on PASCAL VOC 2007 test set. 07+12: union of VOC2007 and VOC2012 trainval.}

\label{tabsln}
\end{table*}

\subsubsection{Hyper-Parameters Selection and Analysis}
Regularization constant $\alpha$ is chosen in such a way that performance loss $\epsilon_1\in [2,3]$. Experimentally, we found that if $\epsilon_1\geq 5$ then at a later stage we can not recover the performance loss due to the pruning of the important connections. Recall that $\Theta_{L1}$ and $\Theta_{L1}^{th}$ denote the model parameters obtained after applying regularization and global thresholding (with threshold $t$) respectively. The threshold $t$ is selected such that there is a further performance drop of $\epsilon_2\in [5,7]$ after thresholding. Note that $\Theta_{L1}^{th}$ is only used for selecting unimportant filters while the actual filter pruning is done on the model with parameters $\Theta_{L1}$. This allows us to afford a relatively high-performance drop after thresholding and explains the high value of $\epsilon_2$. Experimentally, we found that $\epsilon_2\geq 10$ results in too much sparsity in  $\Theta_{L1}^{th}$ and consequently affects the filter selection procedure by selecting important filters for pruning. This results in an irrecoverable loss in performance.

We use grid search for selecting hyperparameters $s_F$, $s_F'$ and $s_G$ in the range $[0.70, 0.99]$ and found $s_F=0.9$, $s_F'=0.85$ and $s_G=0.95$ to work best for $\epsilon_1$ and $\epsilon_2$ in the ranges mentioned above.

\subsubsection{Ablation study on pruned SSD300}

\begin{table}[b]
\centering
\scalebox{1.0}{
\begin{tabular}{|l|l|c|c|c|}
\hline
\textbf{Methods} & \textbf{data} & \textbf{\#Params} & \textbf{Size} & \textbf{mAP} \\ \hline
Pruned-1det & 07+12 & 2.7M & 10.8 MB & 25.6 \\ \hline
Pruned-2det & 07+12 & 3.42M & 13.7 MB & 66.7 \\ \hline
Pruned-3det & 07+12 & 3.7M & 15 MB & 72.15 \\ \hline
Pruned-4det & 07+12 & 3.8M & 15.2 MB & 74.5 \\ \hline
Pruned-5det & 07+12 & 3.85M & 15.4 MB & 74.94 \\ \hline
Pruned-6det & 07+12 & 3.9M & 15.7 MB & 75.07 \\ \hline
\end{tabular}}
 
\caption{Ablation study on pruned SSD300 detection layers on PASCAL VOC07.}

\label{ablatbl}
\end{table}

SSD300-pruned-1det and SSD300-pruned-6det have a significant difference in mAP because PASCAL VOC dataset has high variation in object size. It is clear from Table \ref{ablatbl},  that at least three detection layers are required for a decent mAP. Please note that our pruned models have very fewer parameters in $4^{th}$, $5^{th}$ and $6^{th}$ detection layer but play a significant role in detecting larger objects.


\subsection{Pruning of SSD on GTSDB}\label{secExpGtsdb}

\begin{figure}[t]
\centering
\includegraphics[width=0.48\linewidth]{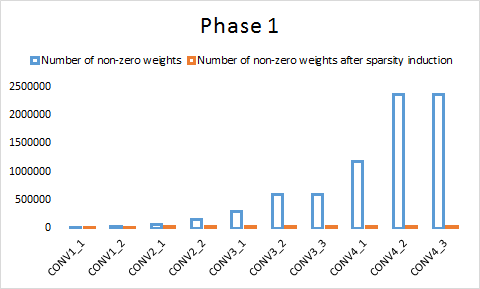}~~~
\includegraphics[width=0.48\linewidth]{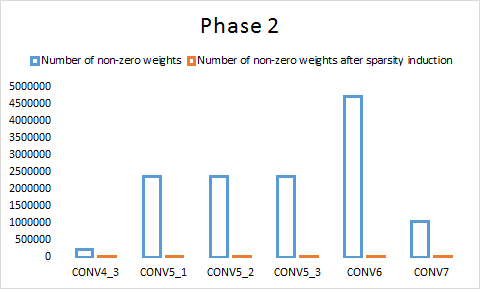}
\caption{Number of non-zero parameters in SSD convolutional layers before and after different phases of our sparsity induction method (GTSDB).}
\label{figssdspg}

\end{figure}

\begin{table*}[t]
\centering
\scalebox{0.60}{
\begin{tabular}{|c|c|c|c|c|c|c|c|c|c|c|c|c|c|c|c|c|c|c|c|c|c|c|c|c|c|}
\hline
 & \multicolumn{25}{c|}{\textbf{CONV Layers}} \\ \hline
\textbf{SSD 512} & \textbf{1\_1} & \textbf{1\_2} & \textbf{2\_1} & \textbf{2\_2} & \textbf{3\_1} & \textbf{3\_2} & \textbf{3\_3} & \textbf{4\_1} & \textbf{4\_2} & \textbf{4\_3} & \textbf{5\_1} & \textbf{5\_2} & \textbf{5\_3} & \textbf{6} & \textbf{7} & \textbf{8\_1} & \textbf{8\_2} & \textbf{9\_1} & \textbf{9\_2} & \textbf{10\_1} & \textbf{10\_2} & \textbf{11\_1} & \textbf{11\_2} & \textbf{12\_1} & \textbf{12\_2} \\ \hline
\textbf{original} & 64 & 64 & 128 & 128 & 256 & 256 & 256 & 512 & 512 & 512 & 512 & 512 & 512 & 1024 & 1024 & 256 & 512 & 128 & 256 & 128 & 256 & 128 & 256 & 128 & 256 \\ \hline
\textbf{pruned-det1} & 64 & 64 & 121 & 123 & 206 & 84 & 113 & 59 & 50 & 189 &  &  &  &  &  &  &  &  &  &  &  &  &  &  &  \\ \hline
\end{tabular}
}
\caption{Layer-wise pruning results in SSD512 model on GTSDB dataset.}

\label{tabssdlayerg}
\end{table*}

\begin{table*}[t]
\centering
\scalebox{0.92}{
\begin{tabular}{|c|c|c|c|c|c|c|}
\hline
\multirow{2}{*}{\textbf{Model}} & \multicolumn{4}{c|}{\textbf{AP}} & \multirow{2}{*}{\textbf{Size}} & \multirow{2}{*}{\textbf{Total Parameters}} \\ \cline{2-5}
 & \textbf{prohibitory} & \textbf{mandatory} & \textbf{danger} & \textbf{mAP} &  &  \\ \hline
\textbf{SSD300-original} & 38.61 & 31.07 & 47.2 & 38.96 & 96.1MB & 24.0M \\ \hline
\textbf{SD512-original} & 88.67 & 76.32 & 82.34 & 82.44 & 98.7 MB & 24.7M \\ \hline
\textbf{SSD512-pruned-1det} & 97.95 & 83.82 & 86.83 & \textbf{89.53} & \textbf{3.8 MB (26X)} & \textbf{938.8K (3.8\%)} \\ \hline
\end{tabular}
}
\caption{AP for each class with original SSD and pruned SSD models on GTSDB.}

\label{tabssdgtsdb}
\end{table*}

We first trained SSD300 and SSD512 on GTSDB and obtained 38.96\% and 82.44\% mAP on test set respectively. The reason for the considerable difference between the performance of SSD300 and SSD512 is because the bounding boxes in GTSDB are very small on average. We only used SSD512 for our pruning experiments. 
For SSD512, we use the same set of layers in each phase as mentioned for the VOC 07+12 dataset in Section~\ref{secExpVoc}. Figure~\ref{figssdspg} shows the number of non-zero parameters in convolutional layers after sparsity induction in first two phases. We observed that most of the parameters in layers after conv4\_3 become zero (see Figure~\ref{figssdspg}). For instance, in conv7 only 3 out of ~1M remain non-zero after thresholding. This is expected as only the early layers in SSD are responsible for detecting of smaller objects. We prune the layers in the first phase and remove all layers after conv4\_3 because of the extremely high sparsity. The resulting model SSD512-pruned-1det which has only one detection layer is 26X smaller than the original model. Table~\ref{tabssdlayerg} shows the layer-wise pruning results. Table~\ref{tabssdgtsdb} shows performance and model size of original SSD and SSD512-pruned-1det.

\subsection{Pruning of VGG16 on CIFAR10 and GTSRB}\label{secExpVgg}

\begin{figure}[t!]
\centering
\scalebox{0.55}{

\includegraphics[width=0.9\linewidth]{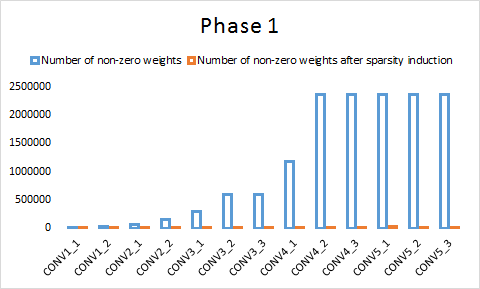}
}
\caption{Number of non-zero parameters in VGG16 convolutional layers before and after sparsity induction on CIFAR10 dataset.}
\label{figvgg}
\end{figure}

 \begin{table}[ht!]
\centering
\scalebox{0.92}{
\begin{tabular}{|c|c|c|c|c|c|}
\hline
\multicolumn{2}{|c|}{\multirow{2}{*}{\textbf{}}} & \multicolumn{2}{c|}{\textbf{VGG 16 on CIFAR-10}} & \multicolumn{2}{c|}{\textbf{VGG 16 on GTSRB}} \\ \cline{3-6} 
\multicolumn{2}{|c|}{} & \textbf{\texttt{Original}} & \textbf{\texttt{Pruned}} & \textbf{\texttt{Original}} & \textbf{\texttt{Pruned}} \\ \hline
\multicolumn{2}{|c|}{\textbf{Input Size}} & \textbf{224x224x3} & \textbf{224x224x3} & \textbf{224x224x3} & \textbf{224x224x3} \\ \hline
\multirow{16}{*}{\textbf{Layers}} & \textbf{3X3 CONV1\_1} & 64 & 43 & 64 & 41 \\ \cline{2-6} 
 & \textbf{3X3 CONV1\_2} & 64 & 24 & 64 & 18 \\ \cline{2-6} 
 & \textbf{3X3 CONV2\_1} & 128 & 53 & 128 & 32 \\ \cline{2-6} 
 & \textbf{3X3 CONV2\_2} & 128 & 43 & 128 & 7 \\ \cline{2-6} 
 & \textbf{3X3 CONV3\_1} & 256 & 58 & 256 & 31 \\ \cline{2-6} 
 & \textbf{3X3 CONV3\_2} & 256 & 60 & 256 & 14 \\ \cline{2-6} 
 & \textbf{3X3 CONV3\_3} & 256 & 68 & 256 & 28 \\ \cline{2-6} 
 & \textbf{3X3 CONV4\_1} & 512 & 97 & 512 & 17 \\ \cline{2-6} 
 & \textbf{3X3 CONV4\_2} & 512 & 104 & 512 & 29 \\ \cline{2-6} 
 & \textbf{3X3 CONV4\_3} & 512 & 121 & 512 & 16 \\ \cline{2-6} 
 & \textbf{3X3 CONV5\_1} & 512 & 127 & 512 & 27 \\ \cline{2-6} 
 & \textbf{3X3 CONV5\_2} & 512 & 55 & 512 & 23 \\ \cline{2-6} 
 & \textbf{3X3 CONV5\_3} & 512 & 113 & 512 & 42 \\ \cline{2-6} 
 & \textbf{FC6} & 4096 & 65 & 4096 & 250 \\ \cline{2-6} 
 & \textbf{FC7} & 4096 & 627 & 4096 & 317 \\ \cline{2-6} 
 & \textbf{FC8} & 10 & 10 & 43 & 43 \\ \hline
\multicolumn{2}{|c|}{\textbf{Total parameters}} & 134.3M & 1.06M (0.8\%) & 134.44M & 663.72K (0.5\%) \\ \hline
\multicolumn{2}{|c|}{\textbf{Model Size}} & 537.2 MB & 4.3 MB (125X) & 537.8 MB & 2.7 MB  (200X)\\ \hline
\multicolumn{2}{|c|}{\textbf{Accuracy}} & 93.23 & 94.01 & 99.31 & 98.76 \\ \hline
\multicolumn{2}{|c|}{\textbf{FLOPs (multiply-adds)}} & 15.47G & 1.47G (9.5\%) & 15.47G & 522.85M (3.4\%) \\ \hline
\end{tabular}
}
\caption{Layer-wise pruning results in VGG16 model on CIFAR10 and GTSRB dataset.}

\label{tabvgglayer}
\end{table}

\begin{table}[ht!]
\centering
\scalebox{1.0}{
\begin{tabular}{|c|c|c|c|}
\hline
\textbf{Model}  & \textbf{Error \%} & \textbf{Params pruned \%} & \textbf{FLOP pruned \%} \\ \hline
\textbf{ICLR'17 \cite{li2016pruning}*} & 6.6 & 64.00\% & 34.20\% \\ \hline
\textbf{NIPS'17 \cite{neklyudov2017structured}} & 7.5 & - & 56.52\% \\ \hline
\textbf{Pruned (ours)} & \textbf{5.99} & \textbf{99.20\%} & \textbf{90.50\%} \\ \hline
\end{tabular}
}
\caption{Comparison of our pruned model (VGG16-Pruned)  with \cite{li2016pruning,neklyudov2017structured} on CIFAR10 dataset. \cite{li2016pruning}* prunes a slightly modified version of VGG16 described in \cite{vgg16mod}.}

\label{tabvgg}
\end{table}

Unlike SSD, VGG16 also has fully connected layers in addition to convolutional layers. We prune the network in two phases. We prune all the convolutional layers in the first phase and the second phase we prune all the fully connected layers. In each phase, we first introduce sparsity in parameters and then prune them as discussed in Section~\ref{secFramework}. For pruning of fully connected layers, we follow Pan et al. \cite{pan2016dropneuron}, where we remove a neuron from a layer if either of the following two conditions is true:\\
\begin{enumerate}
\item all incoming connections' weights are zero
\item all outgoing connections' weights are zero
\end{enumerate}
Figure~\ref{figvgg} shows the number of non-zero parameters in conv layers after sparsity induction.

On CIFAR10, our compressed model is 125X smaller than the original model. Table~\ref{tabvgglayer} shows the layer-wise pruning results and performance of pruned models on CIFAR10 and GTSRB dataset.  

Table~\ref{tabvgg} shows the comparison of our method with the weight sum method by Li et al. \cite{li2016pruning} and the method by Neklyudov et al. \cite{neklyudov2017structured}. Our method prunes 99.2\% of parameters on CIFAR10, significantly larger than 64.0\% pruned by the weight sum method. Furthermore, our method reduces the FLOPs by 90.5\% compared to 34.2\% pruned by the weight sum method.

\section{Conclusion}
In this paper, we have proposed a Multi-layer pruning method. The main idea is to induce sparsity in a first procedure by modifying the training loss function. The second step prunes filter that after being trained with the sparsity-inducing loss-function are zero or nearly zero. This reduced architecture when retrained is observed to provide accurate results for both object detection and classification. Through this method, we have obtained state of the art results for deep model compression for object detection using SSD based object detection on real-world standard benchmark datasets. In the future, we would be interested in exploring this approach for other visual recognition methods and further evaluate other related loss functions that can aid pruning of the filters.
\newpage
{\small
\bibliographystyle{ieee}
\bibliography{egbib}

\begin{thebibliography}{10}\itemsep=-1pt

\bibitem{anisimov2017towards}
D.~Anisimov and T.~Khanova.
\newblock Towards lightweight convolutional neural networks for object
  detection.
\newblock In {\em Advanced Video and Signal Based Surveillance (AVSS), 2017
  14th IEEE International Conference on}, pages 1--8. IEEE, 2017.

\bibitem{Ba2014DoDN}
J.~Ba and R.~Caurana.
\newblock Do deep nets really need to be deep?
\newblock In {\em NIPS}, 2014.

\bibitem{Balan2015BayesianDK}
A.~K. Balan, V.~Rathod, K.~Murphy, and M.~Welling.
\newblock Bayesian dark knowledge.
\newblock In {\em NIPS}, 2015.

\bibitem{bell2016inside}
S.~Bell, C.~Lawrence~Zitnick, K.~Bala, and R.~Girshick.
\newblock Inside-outside net: Detecting objects in context with skip pooling
  and recurrent neural networks.
\newblock In {\em Proceedings of the IEEE conference on computer vision and
  pattern recognition}, pages 2874--2883, 2016.

\bibitem{Bucila2006ModelC}
C.~Bucila, R.~Caruana, and A.~Niculescu-Mizil.
\newblock Model compression.
\newblock In {\em KDD}, 2006.

\bibitem{Chen2017LearningEO}
G.~Chen, W.~Choi, X.~Yu, T.~X. Han, and M.~K. Chandraker.
\newblock Learning efficient object detection models with knowledge
  distillation.
\newblock In {\em NIPS}, 2017.

\bibitem{Chen2015Net2NetAL}
T.~Chen, I.~J. Goodfellow, and J.~Shlens.
\newblock Net2net: Accelerating learning via knowledge transfer.
\newblock {\em CoRR}, abs/1511.05641, 2015.

\bibitem{Chen2015CompressingNN}
W.~Chen, J.~T. Wilson, S.~Tyree, K.~Q. Weinberger, and Y.~Chen.
\newblock Compressing neural networks with the hashing trick.
\newblock In {\em ICML}, 2015.

\bibitem{Cohen2016GroupEC}
T.~Cohen and M.~Welling.
\newblock Group equivariant convolutional networks.
\newblock In {\em ICML}, 2016.

\bibitem{Courbariaux2015BinaryConnectTD}
M.~Courbariaux, Y.~Bengio, and J.-P. David.
\newblock Binaryconnect: Training deep neural networks with binary weights
  during propagations.
\newblock In {\em NIPS}, 2015.

\bibitem{Everingham2014ThePV}
M.~Everingham, S.~M.~A. Eslami, L.~V. Gool, C.~K.~I. Williams, J.~M. Winn, and
  A.~Zisserman.
\newblock The pascal visual object classes challenge: A retrospective.
\newblock {\em International Journal of Computer Vision}, 111:98--136, 2014.

\bibitem{Girshick2015FastR}
R.~B. Girshick.
\newblock Fast r-cnn.
\newblock {\em 2015 IEEE International Conference on Computer Vision (ICCV)},
  pages 1440--1448, 2015.

\bibitem{Han2015DeepCC}
S.~Han, H.~Mao, and W.~J. Dally.
\newblock Deep compression: Compressing deep neural network with pruning,
  trained quantization and huffman coding.
\newblock {\em CoRR}, abs/1510.00149, 2015.

\bibitem{han2016dsd}
S.~Han, J.~Pool, S.~Narang, H.~Mao, S.~Tang, E.~Elsen, B.~Catanzaro, J.~Tran,
  and W.~J. Dally.
\newblock Dsd: Regularizing deep neural networks with dense-sparse-dense
  training flow.
\newblock In {\em ICLR}, 2017.

\bibitem{Hassibi1992SecondOD}
B.~Hassibi and D.~G. Stork.
\newblock Second order derivatives for network pruning: Optimal brain surgeon.
\newblock In {\em NIPS}, 1992.

\bibitem{he2016deep}
K.~He, X.~Zhang, S.~Ren, and J.~Sun.
\newblock Deep residual learning for image recognition.
\newblock In {\em Proceedings of the IEEE conference on computer vision and
  pattern recognition}, pages 770--778, 2016.

\bibitem{he2017channel}
Y.~He, X.~Zhang, and J.~Sun.
\newblock Channel pruning for accelerating very deep neural networks.
\newblock In {\em International Conference on Computer Vision (ICCV)}, 2017.

\bibitem{Hinton2015DistillingTK}
G.~E. Hinton, O.~Vinyals, and J.~Dean.
\newblock Distilling the knowledge in a neural network.
\newblock {\em CoRR}, abs/1503.02531, 2015.

\bibitem{Houben-IJCNN-2013}
S.~Houben, J.~Stallkamp, J.~Salmen, M.~Schlipsing, and C.~Igel.
\newblock Detection of traffic signs in real-world images: The {G}erman
  {T}raffic {S}ign {D}etection {B}enchmark.
\newblock In {\em International Joint Conference on Neural Networks}, number
  1288, 2013.

\bibitem{howard2017mobilenets}
A.~G. Howard, M.~Zhu, B.~Chen, D.~Kalenichenko, W.~Wang, T.~Weyand,
  M.~Andreetto, and H.~Adam.
\newblock Mobilenets: Efficient convolutional neural networks for mobile vision
  applications.
\newblock {\em arXiv preprint arXiv:1704.04861}, 2017.

\bibitem{Hu2016NetworkTA}
H.~Hu, R.~Peng, Y.-W. Tai, and C.-K. Tang.
\newblock Network trimming: A data-driven neuron pruning approach towards
  efficient deep architectures.
\newblock {\em CoRR}, abs/1607.03250, 2016.

\bibitem{NIPS2016_6573}
I.~Hubara, M.~Courbariaux, D.~Soudry, R.~El-Yaniv, and Y.~Bengio.
\newblock Binarized neural networks.
\newblock In D.~D. Lee, M.~Sugiyama, U.~V. Luxburg, I.~Guyon, and R.~Garnett,
  editors, {\em Advances in Neural Information Processing Systems 29}, pages
  4107--4115. Curran Associates, Inc., 2016.

\bibitem{iandola2016squeezenet}
F.~N. Iandola, S.~Han, M.~W. Moskewicz, K.~Ashraf, W.~J. Dally, and K.~Keutzer.
\newblock Squeezenet: Alexnet-level accuracy with 50x fewer parameters and< 0.5
  mb model size.
\newblock {\em arXiv preprint arXiv:1602.07360}, 2016.

\bibitem{kim2016pvanet}
K.-H. Kim, S.~Hong, B.~Roh, Y.~Cheon, and M.~Park.
\newblock Pvanet: deep but lightweight neural networks for real-time object
  detection.
\newblock {\em arXiv preprint arXiv:1608.08021}, 2016.

\bibitem{Lebedev2016FastCU}
V.~Lebedev and V.~S. Lempitsky.
\newblock Fast convnets using group-wise brain damage.
\newblock {\em 2016 IEEE Conference on Computer Vision and Pattern Recognition
  (CVPR)}, pages 2554--2564, 2016.

\bibitem{LeCun1989OptimalBD}
Y.~LeCun, J.~S. Denker, and S.~A. Solla.
\newblock Optimal brain damage.
\newblock In {\em NIPS}, 1989.

\bibitem{Li2016PruningFF}
H.~Li, A.~Kadav, I.~Durdanovic, H.~Samet, and H.~P. Graf.
\newblock Pruning filters for efficient convnets.
\newblock {\em CoRR}, abs/1608.08710, 2016.

\bibitem{li2016pruning}
H.~Li, A.~Kadav, I.~Durdanovic, H.~Samet, and H.~P. Graf.
\newblock Pruning filters for efficient convnets.
\newblock In {\em ICLR}, 2017.

\bibitem{Li2016MultiBiasNA}
H.~Li, W.~Ouyang, and X.~Wang.
\newblock Multi-bias non-linear activation in deep neural networks.
\newblock In {\em ICML}, 2016.

\bibitem{cocomicrosoft}
T.-Y. Lin, M.~Maire, S.~Belongie, J.~Hays, P.~Perona, D.~Ramanan,
  P.~Doll{\'a}r, and C.~L. Zitnick.
\newblock Microsoft coco: Common objects in context.
\newblock In {\em European conference on computer vision}, pages 740--755.
  Springer, 2014.

\bibitem{Liu2016SSDSS}
W.~Liu, D.~Anguelov, D.~Erhan, C.~Szegedy, S.~E. Reed, C.-Y. Fu, and A.~C.
  Berg.
\newblock Ssd: Single shot multibox detector.
\newblock In {\em ECCV}, 2016.

\bibitem{Luo2017ThiNetAF}
J.-H. Luo, J.~Wu, and W.~Lin.
\newblock Thinet: A filter level pruning method for deep neural network
  compression.
\newblock {\em 2017 IEEE International Conference on Computer Vision (ICCV)},
  pages 5068--5076, 2017.

\bibitem{neklyudov2017structured}
K.~Neklyudov, D.~Molchanov, A.~Ashukha, and D.~P. Vetrov.
\newblock Structured bayesian pruning via log-normal multiplicative noise.
\newblock In {\em Advances in Neural Information Processing Systems}, pages
  6778--6787, 2017.

\bibitem{pan2016dropneuron}
W.~Pan, H.~Dong, and Y.~Guo.
\newblock Dropneuron: Simplifying the structure of deep neural networks.
\newblock {\em arXiv preprint arXiv:1606.07326}, 2016.

\bibitem{Rastegari2016XNORNetIC}
M.~Rastegari, V.~Ordonez, J.~Redmon, and A.~Farhadi.
\newblock Xnor-net: Imagenet classification using binary convolutional neural
  networks.
\newblock In {\em ECCV}, 2016.

\bibitem{Shang2016UnderstandingAI}
W.~Shang, K.~Sohn, D.~Almeida, and H.~Lee.
\newblock Understanding and improving convolutional neural networks via
  concatenated rectified linear units.
\newblock In {\em ICML}, 2016.

\bibitem{Srinivas2015DatafreePP}
S.~Srinivas and R.~V. Babu.
\newblock Data-free parameter pruning for deep neural networks.
\newblock In {\em BMVC}, 2015.

\bibitem{Xie2017VisualizationAP}
X.~Xie, X.~Han, Q.~Liao, and G.~Shi.
\newblock Visualization and pruning of ssd with the base network vgg16.
\newblock In {\em ICDLT '17}, 2017.

\bibitem{vgg16mod}
S.~Zagoruyko.
\newblock 92.45\% on cifar-10 in torch.
\newblock \url{http://torch.ch/blog/2015/07/30/cifar.html}, 2015.

\bibitem{Zagoruyko2016PayingMA}
S.~Zagoruyko and N.~Komodakis.
\newblock Paying more attention to attention: Improving the performance of
  convolutional neural networks via attention transfer.
\newblock {\em CoRR}, abs/1612.03928, 2016.

\bibitem{Zhai2016DoublyCN}
S.~Zhai, Y.~Cheng, W.~Lu, and Z.~Zhang.
\newblock Doubly convolutional neural networks.
\newblock In {\em NIPS}, 2016.

\end{thebibliography}


\begin{thebibliography}{1}\itemsep=-1pt

\bibitem{Alpher02}
A.~Alpher.
\newblock Frobnication.
\newblock {\em Journal of Foo}, 12(1):234--778, 2002.

\bibitem{Alpher03}
A.~Alpher and J.~P.~N. Fotheringham-Smythe.
\newblock Frobnication revisited.
\newblock {\em Journal of Foo}, 13(1):234--778, 2003.

\bibitem{Alpher04}
A.~Alpher, J.~P.~N. Fotheringham-Smythe, and G.~Gamow.
\newblock Can a machine frobnicate?
\newblock {\em Journal of Foo}, 14(1):234--778, 2004.

\bibitem{Authors06b}
Authors.
\newblock Frobnication tutorial, 2006.
\newblock Supplied as additional material {\tt tr.pdf}.

\bibitem{Authors06}
Authors.
\newblock The frobnicatable foo filter, 2011.
\newblock Face and Gesture submission ID 324. Supplied as additional material
  {\tt fg324.pdf}.

\end{thebibliography}
}
\section{Appendix}
\subsection{Pruning of SSD on COCO}
We perform additional experiments on the COCO object detection set \cite{cocomicrosoft}.
COCO dataset has 80 object categories. We train SSD300 using \texttt{val35k} images \cite{bell2016inside} (COCO \texttt{val2014} set minus \texttt{minival}) and evaluate on \texttt{minival} (5k images) which is a subset of COCO \texttt{val2014}.

On COCO, our method compresses SSD300 by a factor of $ \sim $ 3X with only a marginal loss in AP.  Table \ref{tabcoco} shows the comparison of our pruned model with the baseline. Table \ref{tabssdlayergcoco} shows the layer-wise filter pruning results in SSD300 model on the COCO dataset.
\begin{table*}[h]
\centering
\begin{tabular}{|c|c|c|c|c|c|c|}
\hline
\multirow{2}{*}{\textbf{Method}} & \multirow{2}{*}{\textbf{Training data}} & \multirow{2}{*}{\textbf{\#Params}} & \multirow{2}{*}{\textbf{\begin{tabular}[c]{@{}c@{}}Model\\ Size\end{tabular}}} & \multicolumn{2}{c|}{\textbf{Avg. Precision, IoU:}} \\ \cline{5-6} 
                                 &                                &                                    &                                                                                & \textbf{0.5:0.95}  & \textbf{0.50}  \\ \hline
SSD300-baseline                  & \texttt{val35k}                         & 35.1M                              & 140.4 MB                                                                       & 21.9               & 38.7                    \\ \hline
SSD300-pruned-6det                    & \texttt{val35k}                         & \textbf{12.7M}                              & \textbf{51.1 MB}                                                                        & 21.7               & \textbf{38.8}                   \\ \hline
\end{tabular}
\caption{Comparison of our pruned model (SSD300-pruned-6det) with the baseline (SSD300) on the COCO \texttt{minival} test set. \texttt{val35k}: 35k images from the subset of the 2014 validation set.}
\label{tabcoco}
\end{table*}

\begin{table*}[h]
\centering
\scalebox{0.68}{
\begin{tabular}{|c|c|c|c|c|c|c|c|c|c|c|c|c|c|c|c|c|c|c|c|c|c|c|c|}
\hline
 & \multicolumn{23}{c|}{\textbf{CONV Layers}} \\ \hline
\textbf{SSD300} & \textbf{1\_1} & \textbf{1\_2} & \textbf{2\_1} & \textbf{2\_2} & \textbf{3\_1} & \textbf{3\_2} & \textbf{3\_3} & \textbf{4\_1} & \textbf{4\_2} & \textbf{4\_3} & \textbf{5\_1} & \textbf{5\_2} & \textbf{5\_3} & \textbf{6} & \textbf{7} & \textbf{8\_1} & \textbf{8\_2} & \textbf{9\_1} & \textbf{9\_2} & \textbf{10\_1} & \textbf{10\_2} & \textbf{11\_1} & \textbf{11\_2}  \\ \hline

\textbf{baseline} & 64 & 64 & 128 & 128 & 256 & 256 & 256 & 512 & 512 & 512 & 512 & 512 & 512 & 1024 & 1024 & 256 & 512 & 128 & 256 & 128 & 256 & 128 & 256  \\ \hline

\textbf{pruned-6det} & 64 & 64 & 125 & 128 & 256 & 255 & 256 & 511 & 365 & 407 & 187 & 222 & 153 & 125 & 276 & 137 & 153 & 95 & 108 & 99 & 81 & 85 & 74  \\ \hline
\end{tabular}
}
\caption{Layer-wise pruning results in SSD300 model on the COCO dataset.}
\label{tabssdlayergcoco}
\end{table*}


\end{document}